\documentclass[conference]{IEEEtran}
\usepackage{times}

\usepackage[numbers]{natbib}
\usepackage{multicol}
\usepackage[bookmarks=true]{hyperref}
\usepackage{cuted}
\usepackage{capt-of}

\usepackage{algorithm}
\usepackage[noend]{algpseudocode}
\usepackage{float}
\usepackage{bbm}
\usepackage{graphicx}
\usepackage{subcaption}
\usepackage{hyperref}       %
\usepackage{url}            %
\usepackage{booktabs}       %
\usepackage{amsfonts}       %
\usepackage{amsmath,amssymb}
\usepackage{nicefrac}       %

\let\labelindent\relax

\usepackage{enumitem}
\usepackage{wrapfig}
\usepackage[table]{xcolor}

\newcommand{\Skip}[1]{}

\newcommand{\ie}{i.e.,\ }

\newcommand{\RNum}[1]{\uppercase\expandafter{\romannumeral #1\relax}}

\newcommand{\ourmethod}{\emph{COMBO-Grasp}}

\begin{document}

\title{COMBO-Grasp: Learning Constraint-Based Manipulation for Bimanual Occluded Grasping}

\author{ 
\authorblockN{Jun Yamada  \qquad
Alexander L. Mitchell \qquad
Jack Collins \qquad
Ingmar Posner \qquad
\authorblockA{University of Oxford}
}}

\maketitle

\begin{strip}
\centering
\includegraphics[width=0.86\textwidth]{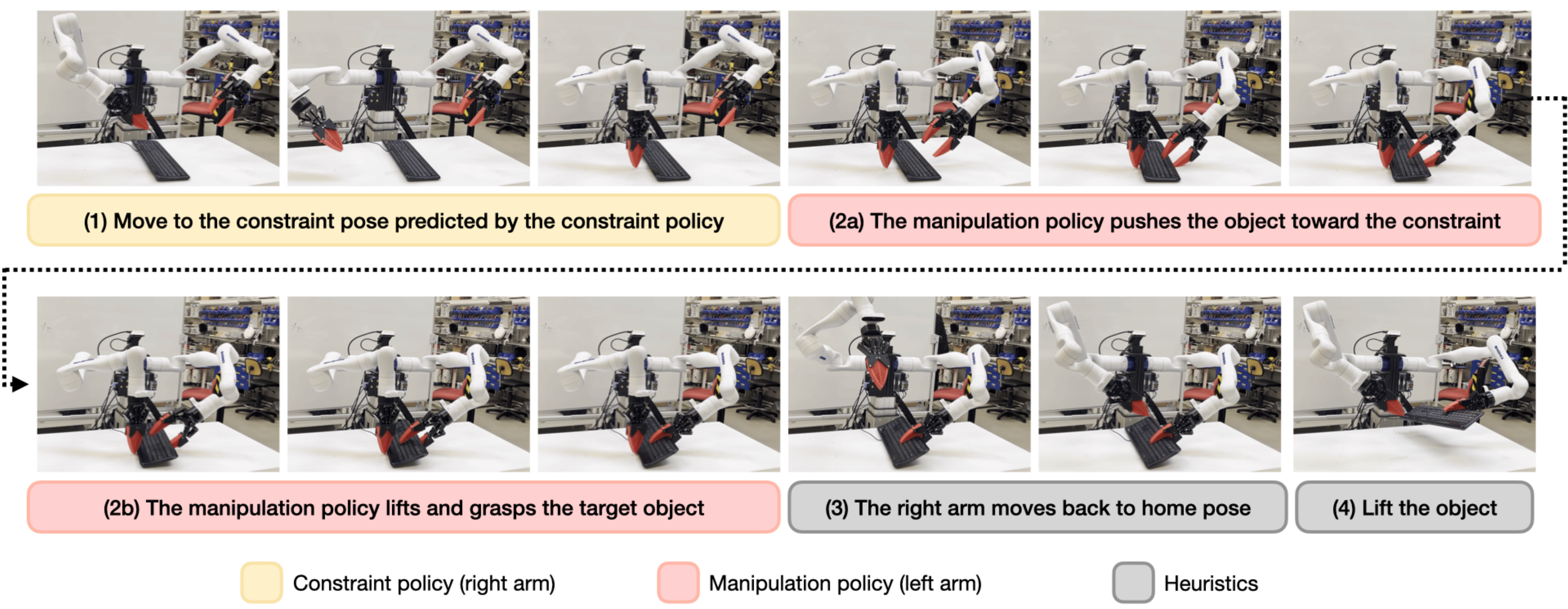}
\captionof{figure}{We introduce \ourmethod, a bimanual robotic system that uses two coordinated policies to address the challenges of grasping objects when the grasp pose is occluded. The system leverages a constraint policy that predicts the pose for the right arm to support the left arm during manipulation. Task execution unfolds in the following sequence: (1) the right arm moves to the predicted support pose using motion planning, (2) the left arm uses the constraint to grasp the target object, (3) the right arm returns to its home position, and (4) the left arm lifts the target object to complete the task.
\label{fig:teaser}}
\vspace{-0.3cm}
\end{strip}

\begin{abstract}
This paper addresses the challenge of \emph{occluded} robot grasping, i.e. grasping in situations where the desired grasp poses are kinematically infeasible due to environmental constraints such as surface collisions.
Traditional robot manipulation approaches struggle with the complexity of non-prehensile or bimanual strategies commonly used by humans in these circumstances.
State-of-the-art reinforcement learning (RL) methods are unsuitable due to the inherent complexity of the task.
In contrast, learning from demonstration requires collecting a significant number of expert demonstrations, which is often infeasible.
Instead, inspired by human bimanual manipulation strategies, where two hands coordinate to stabilise and reorient objects, we focus on a bimanual robotic setup to tackle this challenge.
In particular, we introduce Constraint-based Manipulation for Bimanual Occluded Grasping (\ourmethod), a learning-based approach which leverages two coordinated policies: a constraint policy trained using self-supervised datasets to generate stabilising poses and a grasping policy trained using RL that reorients and grasps the target object.
A key contribution lies in value function-guided policy coordination. 
Specifically, during RL training for the grasping policy, the constraint policy's output is refined through gradients from a jointly trained value function, improving bimanual coordination and task performance.
Lastly, \ourmethod~employs teacher-student policy distillation to effectively deploy point cloud-based policies in real-world environments.
Empirical evaluations demonstrate that \ourmethod~significantly improves task success rates compared to competitive baseline approaches, with successful generalisation to unseen objects in both simulated and real-world environments.

\end{abstract}

\IEEEpeerreviewmaketitle

\section{Introduction}
Grasping objects with kinematically infeasible grasp poses due to environmental collisions, known as occluded grasping~\cite{zhan2022learning},  presents a significant challenge in robotics.
Indeed, kinematic infeasibility arises from supporting surfaces, such as the table that the object is resting on.
For example, grasping a keyboard that rests on a desk requires reorienting the keyboard with regard to the desk surface (nonprehensile manipulation) to reveal the grasp pose (see Figure~\ref{fig:teaser}). 
Humans exhibit exceptional dexterity in solving such occluded grasping problems through coordinated bimanual manipulation, seamlessly using both hands to reposition objects for grasping.
However, learning to acquire such coordinated skills for a bimanual robotic system poses significant challenges, particularly when using reinforcement learning (RL)~\cite{schulman2017proximalpolicyoptimizationalgorithms, haarnoja2018soft}.

Specifically, compared to single-handed applications, bimanual manipulation exhibits a significantly increased action space with coordination requirements adding to task complexity. These challenges are exacerbated when using domain randomisation~\cite{tobin2017domain} to enable sim-to-real transfer and render RL approaches infeasible due to sample inefficiency.
For the occluded grasping task, these challenges are particularly pronounced as the policies must enable one arm to stabilise the object while the other reorients and grasps it.
More importantly, designing a reward function that facilitates the emergence of such coordinated behaviour is nontrivial.
Compared to RL, learning from demonstration (LfD) necessitates a large number of expert demonstrations~\cite{zhao2024alohaunleashedsimplerecipe} encompassing a diverse range of objects to achieve generalisation to unseen objects.

In this work, we present \textbf{Co}nstraint-based \textbf{M}anipulation
for \textbf{B}imanual \textbf{O}ccluded Grasping (\ourmethod), a system specifically designed to address the challenges of occluded grasping using bimanual robot systems.
Drawing inspiration from human bimanual strategies, where the nondominant hand stabilizes an object while the dominant hand performs complex manipulations~\cite{bagesteiro2002handedness, bagesteiro2003, drolet2024comparison}, \ourmethod~consists of two coordinated policies: a \textit{constraint policy} trained using self-supervised learning to generate stabilising poses, and a \textit{grasping policy} trained using RL that reorients and grasps the target object.
The constraint policy generates object-stabilising poses for one of the arms before the grasping policy controls the other arm to attempt grasping.
The use of the constraint policy coupled with the grasping policy accelerates RL training as both policies work in coordination to solve occulded grasping problems in a data-efficient manner.

The constraint policy is trained on a synthetic dataset collected in a self-supervised manner within a simulation designed to leverage force closure as a signal.
At the core of \ourmethod~is value function-guided policy coordination that refines the constraint pose generated by the constraint policy to improve bimanual coordination, thereby enhancing task performance. 
During RL training for the grasping policy, using gradients from the value function trained in tandem with the grasping policy, \ourmethod~optimises the constraint pose to increase the likelihood of a successful grasp once the grasping policy is executed.
This value function-guided policy coordination ensures that the output of the constraint policy is refined to align with the goals of the grasping policy, leading to improved stability of objects during the bimanual grasping.

\ourmethod~achieves effective sim-to-real transfer by adopting a teacher-student policy distillation approach.
The teacher policy, trained with privileged information for diverse objects in simulation, is distilled into a student policy that operates on point clouds for effective sim-to-real transfer. 
In contrast to training a single RL policy or LfD, \ourmethod~learns effective bimanual coordination with improved sampled efficiency and generalizes to unseen objects without relying on any expert demonstrations.

In summary, our contributions are four-fold: 
\begin{itemize}
    \item \ourmethod, a novel approach to bimanual manipulation comprising two coordinated policies to solve occluded grasping problems.
    \item The use of force-closure as a signal to train a self-supervised constraint policy, which accelerates the subsequent RL grasping policy training.
     \item Value function-guided policy coordination that refines generated constraint poses using gradients from the value function to improve coordination during RL training for the grasping policy.
    \item Empirically demonstrating that \ourmethod~successfully grasps seen and unseen objects in both simulated and real-world environments.
\end{itemize}

\section{Related Works}

\subsection{Learning to Grasp Objects}
Grasping is a fundamental robot skill essential for various subsequent manipulation tasks~\cite{yamada2023efficient, collins2023ramp, yuan2023m2t2}.
Many prior works focus on learning grasp pose prediction models with open-loop planning controls~\cite{yuan2023m2t2, mousavian20196, barad2024graspldm}.
These methods typically assume that there exist grasp poses that are not in collision with obstacles, such as a table, so that a robot can reach these poses using motion planning.
Consequently, such open-loop approaches are inadequate for handling occluded grasping tasks where environmental constraints may block or interfere with the target grasp poses.

Prior art has also investigated closed-loop grasping policies using reinforcement learning (RL)~\cite{kalashnikov2018scalable, wang2022goal} and imitation learning (IL)~\cite{zhou2023nerf, Song2019GraspingIT}.
Recent advancements using deep RL have empowered robots to acquire complex manipulation skills driven by predefined reward functions.
For example, \citet{wang2022goal} proposes learning a 6-DoF policy to grasp objects in tabletop environments.
Moreover, QT-Opt~\cite{kalashnikov2018scalable} learns to grasp objects in a box by collecting diverse samples from multiple real-world robots.
\ourmethod~is positioned amongst the works aiming to learn a closed-loop grasp policy, but focuses on more challenging occluded grasping scenarios that necessitate nonprehensile manipulation before grasping.

\subsection{Occluded Grasping}
Several prior work~\cite{sun2020learning, zhou2023learning} attempt to solve occluded grasping tasks through extrinsic dexterity while using only a single arm.
\citet{zhou2023learning} train a policy using RL from state observations in simulation only for a rectangular object to learn extrinsic dexterity via interaction with a wall to reorient the object for grasping.
\citet{sun2020learning} address the challenge of occluded grasping by utilising two arms; however, both arms are primarily employed for object reorientation, and the approach still relies on external constraints, such as a supporting wall.
In contrast to these methods that rely on external constraints, \ourmethod~considers scenarios without external constraints, necessitating the use of one arm to stabilise the object while the other arm re-orients the object.

\subsection{Bimanual Robotic Systems}
Bimanual robotic manipulation has gained increasing attention due to its flexibility and capability to handle complex tasks such as dynamic handovers~\cite{huang2023dynamic}, twisting bottle lids~\cite{lin2024twisting}, and assembly tasks~\cite{drolet2024comparison, shao2020learning}. 
RL~\cite{haarnoja2018soft, schulman2017proximalpolicyoptimizationalgorithms} offers a reward-driven approach to skill acquisition but requires extensive exploration, particularly for high-DoF bimanual tasks. 
Alternatively, learning from demonstrations often demands a large number of expert demonstrations~\cite{zhao2024alohaunleashedsimplerecipe}, which is often costly and impractical for complex bimanual systems, especially in non-prehensile manipulation scenarios.

Several works address these challenges by incorporating inductive biases into RL frameworks. For instance, predefined parameterised skills~\cite{chitnis2020efficient}, intrinsic motivation for efficient exploration~\cite{Chitnis2020Intrinsic}, and symmetry-aware actor-critic networks~\cite{li2023efficient} have shown promise. 
Similarly, \ourmethod~introduces a constraint policy as an inductive bias, specifically tailored for occluded grasping tasks.
In particular, inspired by studies in biopsychology~\cite{Sainburg2001EvidenceFA, bagesteiro2002handedness, bagesteiro2003}, which suggest that the non-dominant arm tends to stabilise while the dominant arm performs more complex movements, \ourmethod~adopts a similar principle. 
One arm is used to stabilise and secure the object, while the other performs non-prehensile manipulation for occluded grasping.

Stabilising an object with one arm to assist the other in manipulation is a well-established strategy. 
For example, \citet{grannen2023stabilize} propose learning two policies: one for stabilising the object’s position and the other for manipulation, applied to a peg-insertion task. 
Similarly, \citet{shao2020learning} adopt a dual-loop learning framework, where the inner loop optimises the grasping policy, and the outer loop refines the constraint policy based on the inner-loop performance. This approach, though effective, suffers from significant sample inefficiency.

In contrast, \ourmethod~eliminates the reliance on expert demonstrations or dual-loop training. Instead, it utilises self-supervised data collection in simulation to train a constraint policy that stabilises objects, thereby accelerating RL training of the grasping policy for occluded grasping of diverse objects. 
Crucially, \ourmethod~introduces value function-guided policy coordination to refine the generated constraint poses by leveraging gradients derived from maximising the value function of the grasping policy during the RL training. 
This refinement process enables the constraint policy to adapt the constraint pose such that it is more suitable for the grasping policy, thereby improving the coordination between the constraint policy and the grasping policy for bimanual occluded grasping tasks.
\section{Task and System Setup}
\textbf{Task description.} 
\label{sec:task_desc}
This work tackles bimanual occluded grasping problems.
Occluded grasping refers to the inability to execute a desired grasp pose due to collisions between the robot and its environment.

In order to grasp a target object given a desired grasp pose that is occluded, one arm is needed to prevent the object from moving, while the dominant arm attempts to reorient and grasp the object.
In this work, the left robot arm (dominant arm) always attempts to grasp a target object while the right arm (non-dominant arm) stabilises the object to assist the left arm.
We leave dynamic role assignment of left and right arms to future work, similar to~\cite{grannen2023stabilize}.
Moreover, the gripper of the left arm autonomously closes at the end of each episode to grasp the target object, and the left end-effector moves upward to lift the object.

We formulate two distinct Markov Decision Processes (MDP) for the occluded grasping task.
The first MDP is fully observable and defined by a tuple $\mathcal{M}_{1} = (\mathcal{S}_{1}, \mathcal{A}, \mathcal{R}, \gamma, \mathcal{P}_{1})$, where $\mathcal{S}$ is the state space, $\mathcal{A}$ is the action space, $\mathcal{R}$ is the reward function, $\gamma$ is a discount factor, and $\mathcal{P}$ is the environment dynamics.
The second MDP is partially observable, a POMDP, and defined by a tuple $\mathcal{M}_{2} = (\mathcal{S}_{2}, \mathcal{O}_{2}, \mathcal{A}, \mathcal{R}, \gamma, \mathcal{P}_{2})$, where $\mathcal{O}$ is the observation space which includes point clouds.
The state-based teacher constraint and grasping policies are trained in simulation under the MDP $\mathcal{M}_{1}$.
On the other hand, the vision-based student constraint and grasping policies are evaluated in the simulated and real-world environments under the MDP $\mathcal{M}_{2}$.
The RL policy is trained to maximise the cumulative reward $\sum_{t=0}^{T-1}\gamma^{t} \mathcal{r}(\mathbf{s}_{t}, \mathbf{a}_{t})$ where $r \in \mathcal{R}$.

\begin{figure}
    \centering
    \includegraphics[width=0.7\linewidth]{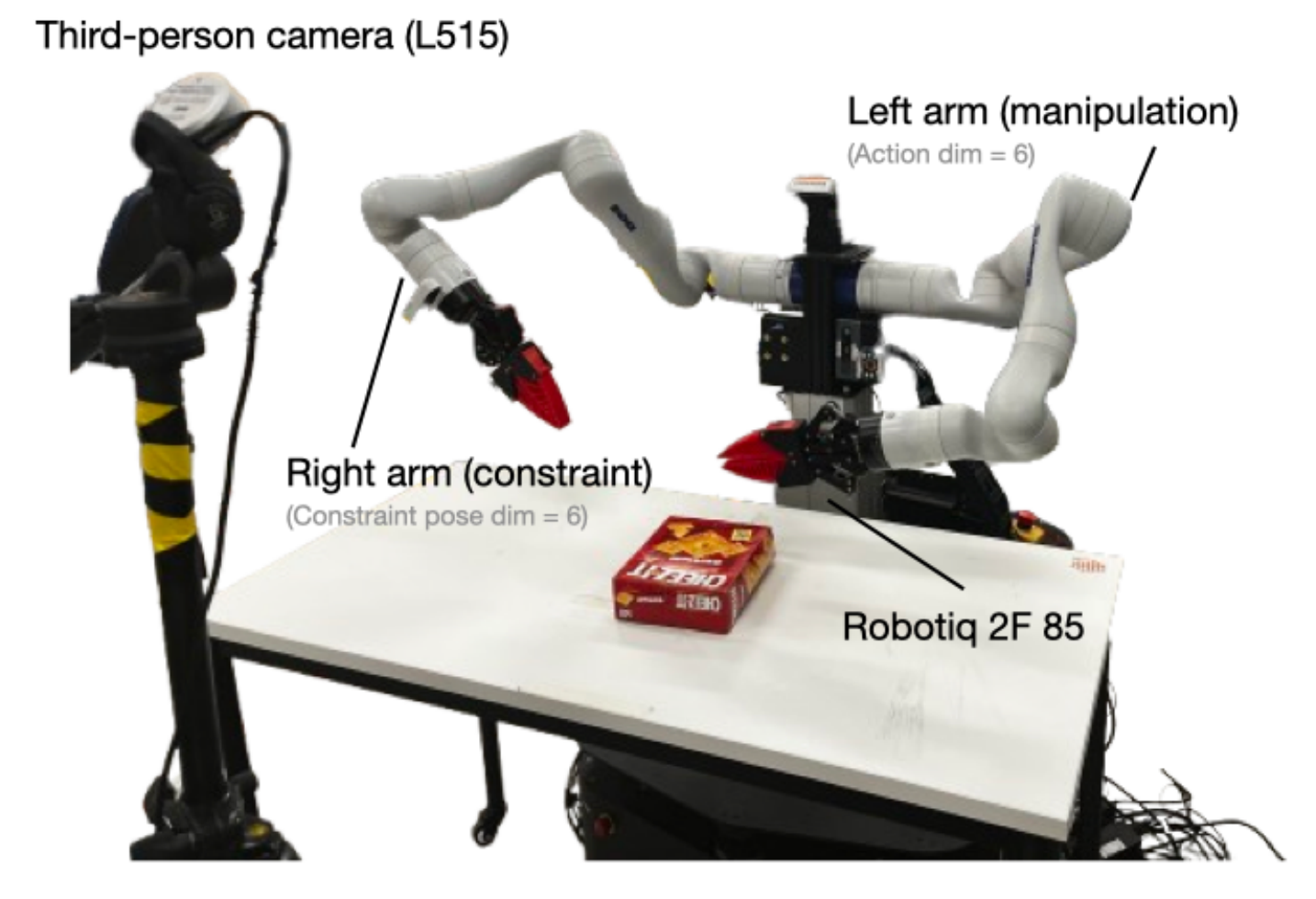}
    \caption{\textbf{Real-world system setup.}  The system comprises two Kinova Gen3 robotic arms mounted perpendicularly to the main body. Each arm is equipped with a Robotiq 2F-85 gripper. To enhance grasping performance, the grippers are fitted with soft fingertips~\cite{chi2024universalmanipulationinterfaceinthewild} instead of the standard ones. Visual observations are captured using a third-person RealSense L515 camera positioned in front of the robot.}
    \label{fig:real-world_sys}
    \vspace{-0.5cm}
\end{figure}

\textbf{Action Space}
The teacher and student policies share the same action space.
A grasping policy that controls the left arm outputs six-dimensional delta poses, including translations and rotations represented in axis-angle format.
As described in Section~\ref{sec:task_desc}, the robot autonomously closes the left gripper at the end of the episode.
Thus, the gasping policy does not include a discrete action for closing and opening the gripper.
A constraint policy for the right arm produces a six-dimensional absolute pose for the right end-effector.
As in prior work~\cite{shao2020learning}, the constraint policy focuses on the x-y plane, given that the constrained end-effector is typically positioned on the table.
Thus, the first two dimensions correspond to the $x$ and $y$ coordinates of the end-effector, while the remaining four define its orientation as a quaternion.
For the baseline RL policy that jointly controls both arms, the policy outputs 12-dimensional delta poses. 
The first six dimensions correspond to the delta actions for the left arm, while the remaining six dimensions are allocated to the right arm.

\textbf{Real-World Setup.}
We design a system for bimanual occluded grasping as shown in Fig.~\ref{fig:real-world_sys}.
The system consists of two Kinova Gen3 arms with Robotq 2F-85 grippers.
The robot arms are perpendicularly attached to a body.
Instead of the original rigid fingertips attached to the Robotiq 2F-85 grippers, deformable fingertips~\cite{chi2024universal} designed for a more secure grip are used.
An L515 Realsense camera, with known extrinsic parameters, serves as the third-person camera and provides point clouds to vision-based student policies.
To control a robot, we use a hybrid task and joint space impedance controller~\cite{kim2019model}.

\textbf{Simulation Setup.}
Isaac Sim~\cite{IsaacDeveloper} is used to train teacher policies for the occluded grasping task.
To train policies, $48$ objects selected from the Google Scanned Objects dataset~\cite{downs2022googlescannedobjectshighquality} are spawned into the environment (see Figure~\ref{fig:training_objects}).
The trained teacher policies are distilled to student policies that take as input point cloud observations obtained from a third-person camera.
We use an operational space controller~\cite{khatib1987unified} to control robot arms.
Further information regarding the simulation setup can be found in Appendix~\ref{appendix:teacher}.

\begin{figure*}
    \centering
    \includegraphics[width=0.85\textwidth]{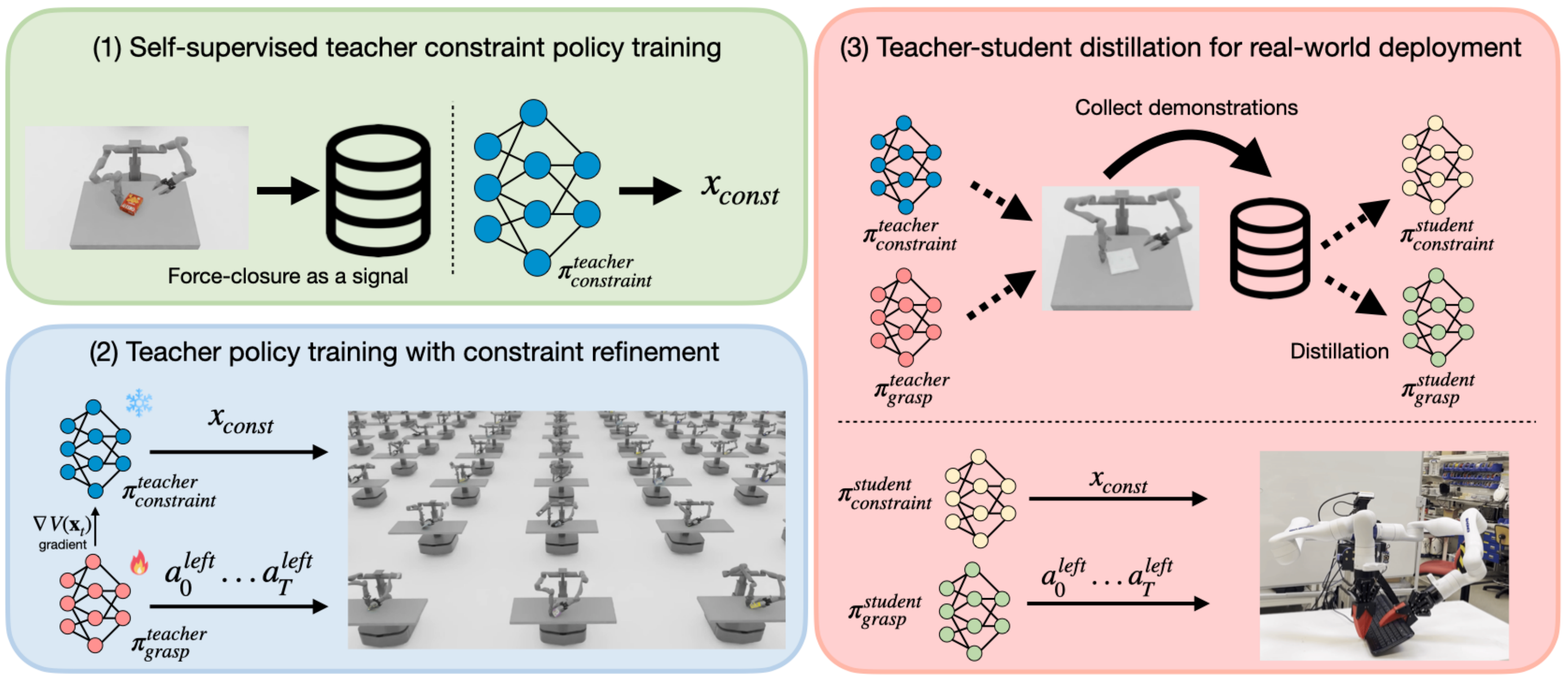}
    \caption{\textbf{Method Overview.} (1) \ourmethod~first collects a synthetic dataset in a self-supervised manner in simulation to train the state-based teacher constraint policy. The teacher constraint policy outputs an end-effector pose for the right arm, given the privileged information available in the simulation. (2) The weights of the trained teacher constraint policy are frozen, and a teacher grasping policy, $\pi_{teacher}$, is trained using RL from privileged information in simulation. To maximise the performance, we propose value function-guided policy coordination that refines the output of the constraint policy using gradients propagated from the value function that is jointly trained with the grasping policy by maximising its value. (3) The teacher grasping policy and the teacher constraint policy are distilled into vision-based student policies that leverage point cloud observations, robot proprioceptive states, and, optionally, a desired grasp pose to address bimanual occluded grasping tasks in real-world environments.}
    \label{fig:pipeline}
    \vspace{-0.5cm}
\end{figure*}

\section{Approach}
In this section we present \ourmethod, a system designed to solve challenging bimanual occluded grasping tasks.
Bimanual occluded grasping refers to scenarios where a desired grasp pose is initially occluded and in collision with external environments such as a table.
COMBO-Grasp utilises two coordinated policies: a constraint policy trained on a dataset collected without human supervision within a simulation to stabilise the target object using one arm, and a grasping policy trained using RL to control the other arm and reorient the object for successful grasping.

This section first introduces a novel self-supervised data collection method (Section~\ref{sec:data_collection}) that creates the data in simulation to train the teacher constraint policy (Section~\ref{sec:pretraining}). 
The training procedure for the teacher grasping policy is then described in Section~\ref{sec:teacher_training}. 
Value function-guided policy coordination that refines the generated constraint pose during RL training for the teacher grasping policy is described in Section~\ref{sec:teacher_training}. Finally, the process of teacher-student distillation for real-world deployment is detailed in Section~\ref{sec:distillation}.

\subsection{Self-Supervised Data Collection for Constraint Policy}
\label{sec:data_collection}

Rather than relying on costly expert demonstrations for training, this work introduces a novel self-supervised data collection approach that uses force-closure as a signal in simulation to train the constraint policy across a diverse set of objects (see Figure~\ref{fig:pipeline} (1)).
Firstly, target occluded grasp pose annotations for objects are generated using antipodal sampling~\cite{eppner2019billion}.
The desired grasp poses are collected for $48$ objects selected from the Google Scanned Objects dataset~\cite{downs2022googlescannedobjectshighquality} (see Figure~\ref{fig:training_objects} in Appendix~\ref{appendix:teacher}).
These desired grasp poses are also used during RL training for a grasping policy (see Section~\ref{sec:teacher_training}).

End-effector poses for the right arm are randomly sampled near the target object placed on a table, while the left arm remains fixed in its initial position.
A force of $25N \times \text{mass}$ along the approach vector of a desired grasp pose is applied to the object.
To assess force closure, the object's velocity is used as an approximation, as accurately evaluating force closure is often challenging, particularly in scenarios involving multiple contacts.
Instead, force closure is considered successful if, after applying force, the object's velocity remains below a predefined threshold, given the specific grasp and constraint poses.
The sampled end-effector pose, the corresponding desired grasp pose, and the object pose are then added to the dataset. 
By iterating this process in the simulation, $3K$ constraint poses per object are collected.
With $48$ objects, this results in a total of $144K$ samples.
By leveraging the object's motion as a proxy measure for the success of a constraint pose, we can generate a rich set of training data to train the constraint policy.

\subsection{Teacher Constraint Policy Training }
\label{sec:pretraining}
One of the central contributions of this work lies in value function-guided policy coordination that integrates classifier guidance in diffusion models to refine the generated constraint pose during the training of the state-based teacher grasping policy. 
This is achieved by employing a diffusion model for the state-based teacher constraint policy, denoted as $\pi^{teacher}_{const}$, trained from the privileged information in the dataset (see Section~\ref{sec:data_collection}).
This approach leverages gradients from the value function to steer the teacher constraint policy's output, optimising the stabilising poses to align with the grasping policy's objectives to improve task performance and sample efficiency.

The teacher constraint policy uses a diffusion model formulated as a Denoising Diffusion Probabilistic Model (DDPM)~\cite{ho2020denoising}.
DDPMs are a class of generative models where the output generation is modelled as a denoising process.
Starting from $x^{K}$ sampled from Gaussian noise, the DDPM performs $K$ denoising iterations to generate a series of intermediate samples with decreasing levels of noise, $x^{k}, x^{k-1},...x^{0}$.
This process is formulated as
\begin{equation}
    \mathbf{x}^{k-1} = \alpha(\mathbf{x}^{k} - \gamma \epsilon_{\theta}(\mathbf{x}^{k}, k) + \mathcal{N}(0, \sigma^2I))
\end{equation}
where $\epsilon_{\theta}$ is a noise prediction network parameterised by $\theta$.
The choice of $\alpha, \gamma, \sigma$ is determined by a noise scheduler.
To train the constraint policy, a forward diffusion process is applied to add noise to an unmodified sample, $x^{0}$, from the dataset by randomly sampling a denoising iteration $k$ and random noise $\epsilon^{k}$.
The noise prediction model $\epsilon_{\theta}$ is then trained to estimate the noise added to a sample during the forward diffusion process.
Thus, the training loss is formulated as
\begin{equation}
    \mathcal{L}_{constraint} = \textit{MSE}(\epsilon^{k}, \epsilon_{\theta}(\mathbf{x}^{0}_{const} + \epsilon^{k}, k))
\end{equation}
where $\mathbf{x}_{const}$ is the constraint pose for the right arm.
In this work, we employ an MLP-based denoising model as the backbone for the diffusion policy (see Appendix~\ref{appendix:teacher_const} for further details of the architecture).

The constraint policy takes as input an object pose, a desired grasp pose, and object IDs.
For the Object IDs, we train an autoencoder~\cite{rumelhart1986} by reconstructing point clouds of objects using the Chamfer distance to learn compact representations similar to prior work~\cite{wan2023unidexgraspimprovingdexterousgrasping}.
Then, we use the learnt compact latent representation as the Object ID rather than using a one-hot vector, as this reduces the dimensionality of the observation space, especially when considering greater numbers of objects.

The state-based teacher constraint policy is used exclusively during the teacher grasping policy training phase, as described in Section~\ref{sec:teacher_training}. 
At a later stage, the teacher constraint policy is distilled into a vision-based student constraint policy for sim-to-real transfer, ensuring robust performance across deployment scenarios.

\subsection{Teacher Grasping Policy}
\label{sec:teacher_training}
After the constraint policy is trained, a teacher grasping policy $\pi^{teacher}_{grasp}$ is trained using Proximal Policy Optimisation (PPO)~\cite{schulman2017proximalpolicyoptimizationalgorithms} for diverse objects from privileged information in simulation.
To train a robust teacher grasping policy capable of performing in real-world environments, we employ domain randomisation, incorporating additive Gaussian noise into low-dimensional observations, as well as randomising the physics parameters of the target object and the controller parameters during policy training. For further information about the domain randomisation, see Appendix~\ref{appendix:sim_training}.
The teacher grasping policy receives as input the robot's proprioceptive states, object pose, object velocity, desired grasp poses, object IDs (see Section~\ref{sec:pretraining}), object's mass and friction parameters, and the PID gains for the OSC controller.

At the beginning of each training episode, the teacher constraint policy $\pi^{teacher}_{const}$ generates a constraint end-effector pose $\mathbf{x}_{const}$ for the right arm.
Given the constraint end-effector pose, the joint positions of the right arm are computed using an inverse kinematics solver in CuRobo~\cite{sundaralingam2023curobo}.
Then, the right arm moves to the computed desired constraint joint positions.
Once the right arm is positioned, the grasping policy controls the left arm to attempt the occluded grasping task.

In this work, we design a reward function using six components: (1) distance reward between a left end-effector position and a desired grasp position, (2) distance reward between the left end-effector orientation and the desired grasp orientation, (3) action penalty to penalise the prediction of large actions by the grasping policy, (4) collision penalty including both self-collision and collision between robot arms and the table, (5) lift reward for incentivising the left arm to reorient the target object to expose the occluded target grasp pose, and (6) a sparse grasp success reward. 
The collision penalty term is computed using the sign distance field provided by CuRobo.
The final reward $r$ is
\begin{equation}
    \begin{split}
    r = \alpha_{1} r_{dist\_pos} &+ 
    \alpha_{2} r_{dist\_ori} - \alpha_{3} r_{collision} \\
    & - \alpha_{4} r_{action} + \alpha_{5} r_{lift} + \alpha_{6} r_{success}        
    \end{split}
\end{equation}
where $\alpha_{i}$ is a coefficient for each reward term.
For more details on teacher policy training, domain randomisation, and each reward term with the coefficient value, see Appendix~\ref{appendix:teacher}.

\subsection{Value Function-guided Policy Coordination}
\label{sec:steering}
A key aspect of \ourmethod~is to induce effective bimanual coordination using the trained constraint policy, thereby improving task performance and enhancing the sample efficiency of RL policy training. 
Since the teacher constraint policy is initially trained on datasets collected using force closure as a signal, it does not inherently guarantee the generation of an optimal constraint for the grasping policy. 
To address this limitation, \ourmethod~draws inspiration from classifier guidance in diffusion models and propose value function-guided policy coordination that refines the generated constraint pose using gradients from a value function $V(\mathbf{x}_{t})$, which is trained alongside the grasping policy using RL.
The value function of the grasping policy acts as a classifier in the classifier guidance framework, and the gradients for guidance are obtained by maximising the estimated value.
This approach effectively refines the generated constraint poses to align more closely with the grasping policy's requirements, leading to improved overall performance and sample efficiency.

By incorporating gradients from the value function by maximisation, the denoising process for the constraint policy is formulated as

\begin{equation}
\label{eq:guidance}
    \mathbf{x}^{k-1}_{const} = \alpha(\mathbf{x}^{k}_{const} - \gamma \epsilon_{\theta}(\mathbf{x}^{k}_{const}, k) - w\nabla V(\mathbf{x}) + \mathcal{N}(0, \sigma^2I))
\end{equation}
where $w$ is a scaling parameter, $\mathbf{x}$ is low-dimensional observation used as input to the value function $V(\cdot)$, and the constraint pose $\mathbf{x}_{const}$ is a subset of the input state $\mathbf{x}$ for the value function (\ie $\mathbf{x}_{const} \in \mathbf{x}$).
For further details on value function-guided policy coordination, see Appendix~\ref{appendix:teacher}.

\subsection{Policy Distillation for Sim-to-Real Transfer}
\label{sec:distillation}

\begin{figure}
    \centering
    \includegraphics[width=0.85\linewidth]{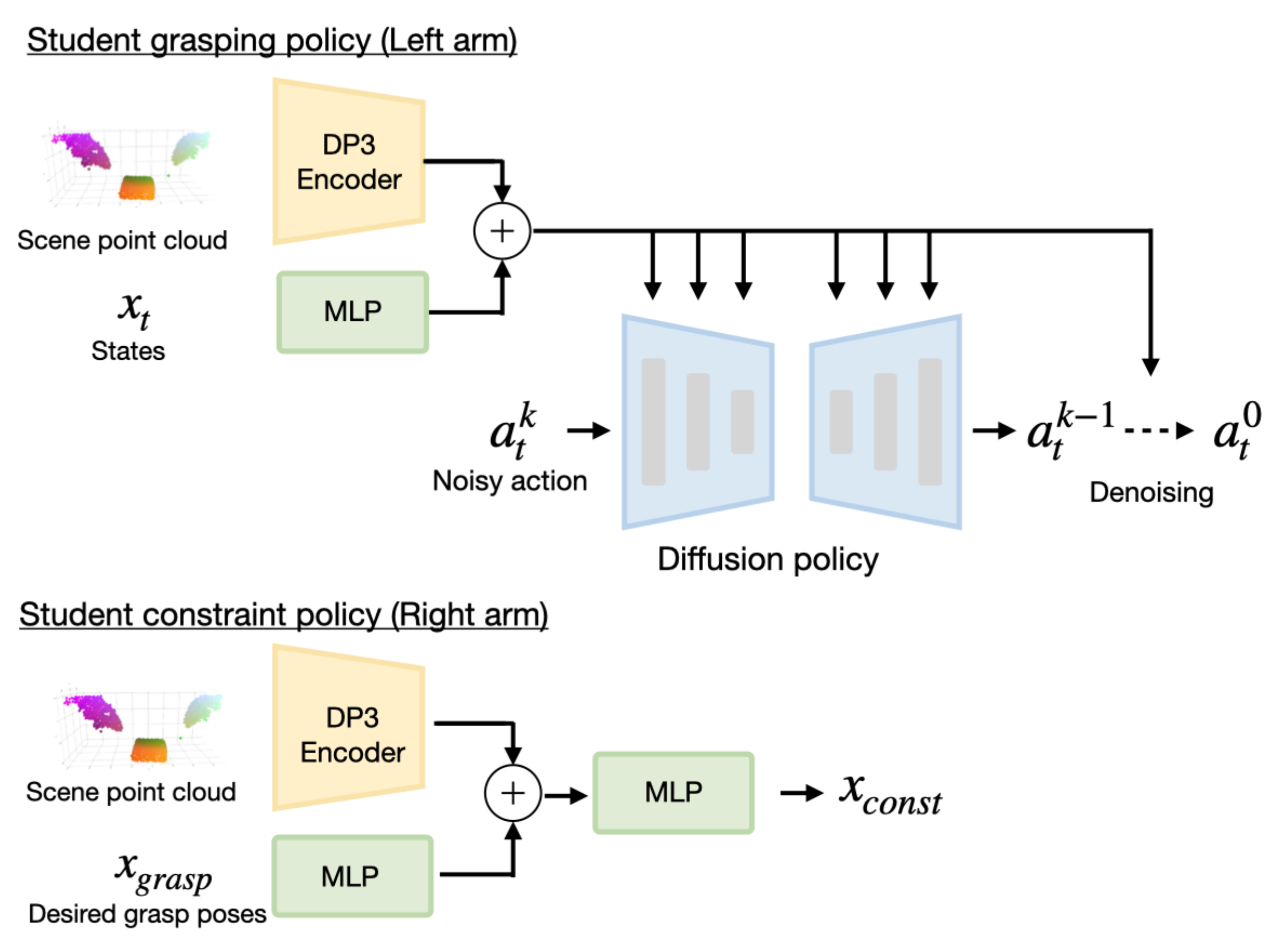}
    \caption{\textbf{Student policy architecture.} We utilize DP3~\cite{ze20243d} as the backbone for the grasping policy. The DP3 encoder processes the scene point cloud, and its output is concatenated with a state feature vector obtained by a multi-layer perceptron (MLP). The resulting concatenated vector serves as the conditioning input for the diffusion-based policy. Similarly, the constraint student policy employs the DP3 encoder and an MLP, but it takes a desired grasp pose as input. Unlike the grasping policy, the constraint student policy employs a Gaussian Mixture Model (GMM)-based policy.}
    \label{fig:student_arch}
    \vspace{-0.3cm}
\end{figure}

To deploy a policy in real-world environments and grasp unseen objects, it is essential to leverage visual observations as an input modality to the policy. 
To achieve this, teacher-student policy distillation~\cite{yamada2024twist, brosseit2021distilled} is used to transfer knowledge from the learnt teacher constraint and grasping policies to student policies. 
These student policies are designed to process point cloud observations and state observations, such as proprioceptive information and, optionally, a desired grasp pose. 
While the inclusion of the desired grasp pose slightly improves performance by biasing the arm's movements toward the target, omitting them still enables effective execution, albeit with a minor performance drop.
In \ourmethod, we employ a diffusion policy as the student grasping policy, similar to that used in the prior work~\cite{ze20243d}.
The student policy consists of DP3~\cite{ze20243d} and MLP encoders to process point cloud and state observations, as shown in Figure~\ref{fig:student_arch}.
The resulting vectors from these encoders are concatenated and used as the conditioning input for the diffusion policy.
In contrast to the student grasping policy, the student constraint policy uses a Gaussian Mixture Model (GMM)-based policy for its simplicity. 
Since the student constraint policy does not require output steering like the teacher policy, the GMM approach is both effective and straightforward.

To distil the teacher to the student policy, we rollout the teacher policy in the simulation and collect expert demonstrations.
The expert demonstrations include visual observations.
We collect $10K$ expert demonstrations for use in the distillation process.
During distillation, we apply a small perturbation to point cloud observations to simulate noise seen in real-world scenes.
For further details on the student policy training, see Appendix~\ref{appendix:student}.

\section{Experimental Results: Simulation}

Our experimental evaluation aims to address the following questions: (1) How successful is \ourmethod~in learning a teacher policy compared to competitive baselines? (2) How well does \ourmethod~generalise to unseen objects? 
(3) How does the value function-guided policy coordination affect \ourmethod's overall performance?
For further analysis of the experiment, see Appendix~\ref{appendix:analysis}.

\subsection{Baselines}
We compare \ourmethod~with the following baselines:

\begin{itemize}
    \item \textbf{PPO}: A PPO~\cite{schulman2017proximal} policy that controls both arms. The policy outputs $12$-dimensional actions. Compared to \ourmethod~which employs two coordinated policies, this baseline requires more extensive exploration to solve the task.
    \item \textbf{PPO + Constraint Reward}: A PPO policy trained using a reward function that adds a distance-based reward between the right end-effector and the centre of the target object into the original reward function. This modification encourages the right arm to act as a constraint, assisting the left arm in grasping tasks. This enables the policy to avoid undesirable physical behaviours observed when using the original reward function, such as the left end-effector attempting to grasp the target object with high velocity without using the constraint.
    \item \textbf{\ourmethod~with a fixed constraint}: A PPO policy is trained to control the left arm, while the right arm remains fixed in a predefined pose in contrast to \ourmethod. This showcases the importance of a constraint policy.
    \item \textbf{\ourmethod~without refinmenet}: \ourmethod~without value function-guided policy coordination. This demonstrates the necessity of refining the constraint pose generated by the constraint policy to further improve performance. 
\end{itemize}

\subsection{Evaluation Metric}
For evaluation, we assess the success rate of grasping. 
In particular, a trial is considered successful if the robot's left arm securely grasps and lifts the target object at least $8 \ cm$ at the end of the episode.

\subsection{Sample Efficiency in Teacher Policy Training}
\begin{figure}
    \centering
    \includegraphics[width=0.75\linewidth]{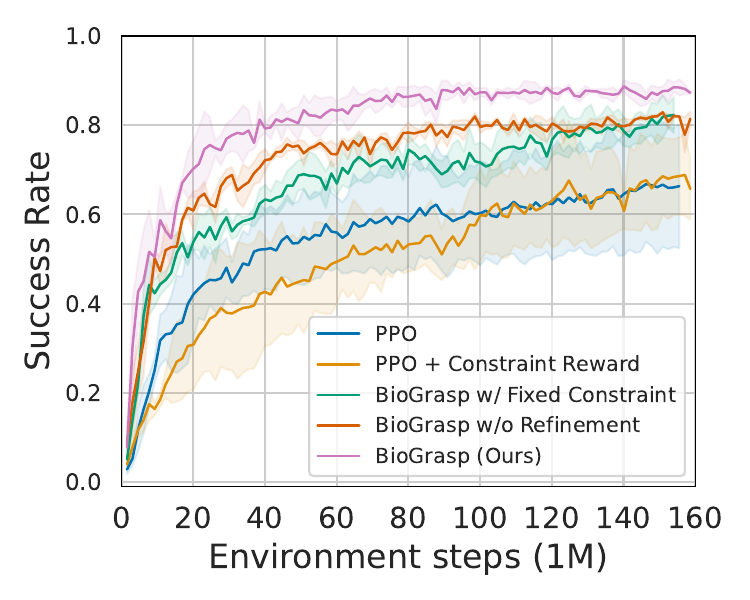}
    \caption{\textbf{Teacher policy training.} We run $3$ seeds for each method, and the shaded region represents the standard deviation. \ourmethod~significantly outperforms competitive baselines in both performance and sample efficiency.}
    \label{fig:teacher_policy}
\end{figure}

Firstly, we evaluate the performance of teacher policy training in simulation.
As shown in Figure~\ref{fig:teacher_policy}, \ourmethod~solves the complex occluded grasping task with more sample efficiency and overall achieves a higher performance. 
On the other hand, \emph{PPO} struggles to achieve similar performance due to the task and system complexity.
More crucially, \emph{PPO} trained with the original reward function often demonstrates undesirable physical behaviours, such that the left arm attempts to grasp the target object with high velocity without leveraging the right arm as a constraint.
Such behaviour exploits the imperfect physics in the simulator and cannot be transferred to real-world environments.
\emph{PPO + Constraint Reward} alleviates this issue by adding a distance-based reward function.
However, defining a reward function to induce a suitable constraint is challenging because such constraint poses are not known beforehand as this is a cooperative task dependent on the approach taken by both arms.
Thus, engineering an appropriate reward function to elicit desired behaviour is not straightforward.
These findings suggest that coordinated policies consisting of the constraint policy and the grasping policy effectively accelerate training and lead to higher success rates compared to a single policy trained using the state-of-the-art RL method.

\ourmethod~w/ fixed constraints demonstrate mediocre performance, as fixed constraints are sometimes not ideal for the grasping policy to solve occluded grasping tasks.
Similarly, \ourmethod~w/o refinement shows worse performance compared to that of \ourmethod.
These results indicate that pre-training a constraint policy and refining the output during the teacher policy training are essential components in \ourmethod.

\subsection{Student Policy Performance in Simulation}
\begin{figure}[t]
    \centering
    \includegraphics[width=0.85\linewidth]{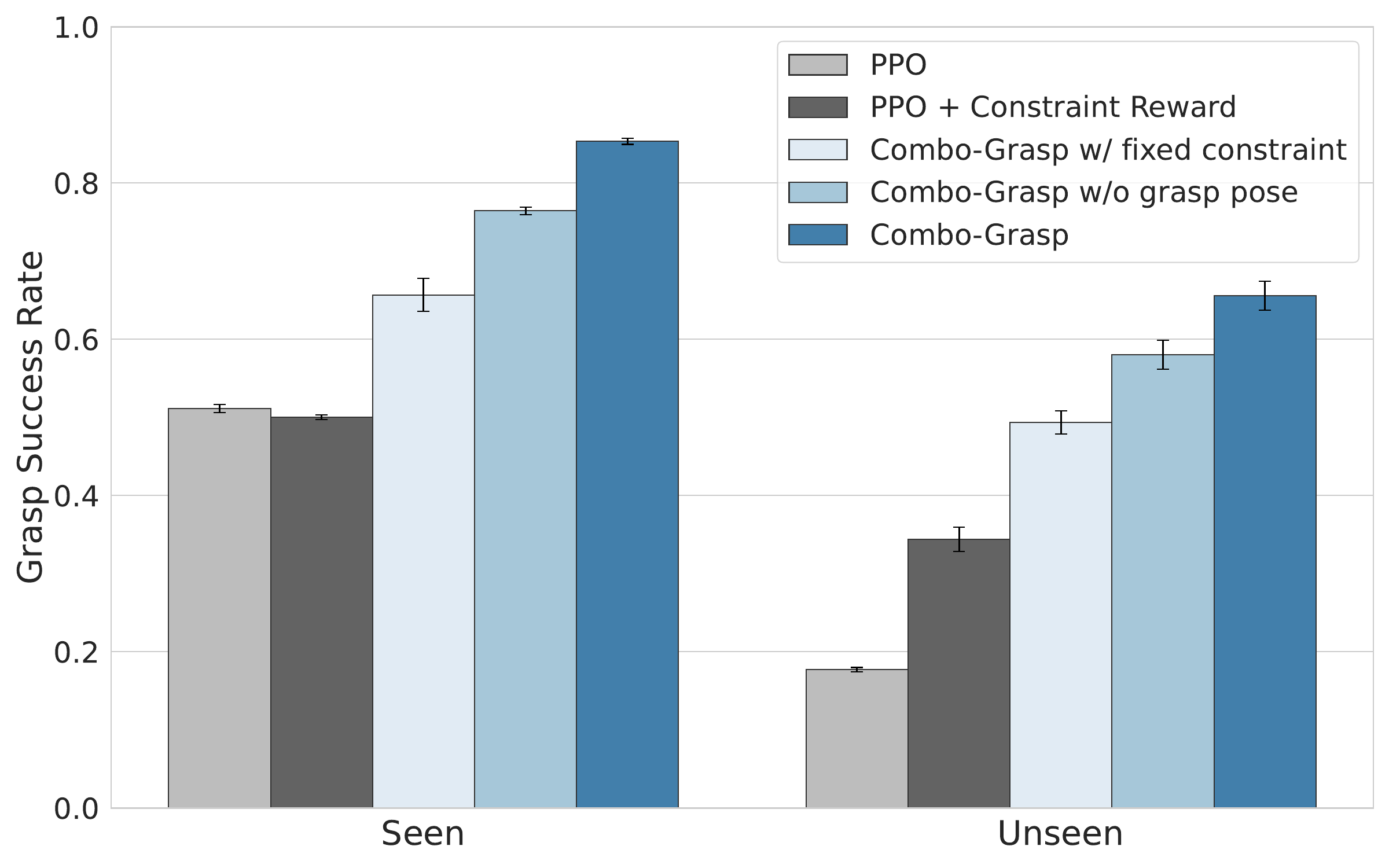}
    \caption{Student Policy Performance averaged over $3$ seeds in Simulated environments. We evaluate each approach for $50$ times using both seen and unseen objects.}
    \label{fig:student_perform}
    \vspace{-0.4cm}
\end{figure}

We assess the performance of the distilled student policies on both seen and unseen objects in the simulation.
Figure~\ref{fig:student_perform} shows the performance of the student policies in the simulated environments.
\ourmethod~effectively addresses the challenging occluded grasping tasks for both seen and unseen objects.
Without the desired grasp pose as input, \ourmethod~has reduced performance; however, it still demonstrates relatively performant results.

The success rates of \emph{PPO} and \emph{PPO + Constraint Reward} are similar for seen objects, but \emph{PPO + Constraint Reward} performs significantly better on unseen objects.
This improvement arises because \emph{PPO + Constraint Reward} avoids exploiting imperfect physics and instead uses the right arm as a constraint induced by the constraint reward, which makes it more transferable to unseen objects. 
These findings suggest that learning a coordinated strategy is crucial for better transfer performance when dealing with unseen objects. 
Nonetheless, the performance of \emph{PPO + Constraint Reward} remains inferior to \ourmethod~due to the difficulty of training a teacher policy that effectively learns a better constraint. 
The challenge of reward engineering limits the teacher's capability, which in turn reduces the performance of the student policy.

\subsection{Ablation of the Value Function-guided Policy Coordination}

\begin{figure}[t]
    \centering
    \includegraphics[width=0.7\linewidth]{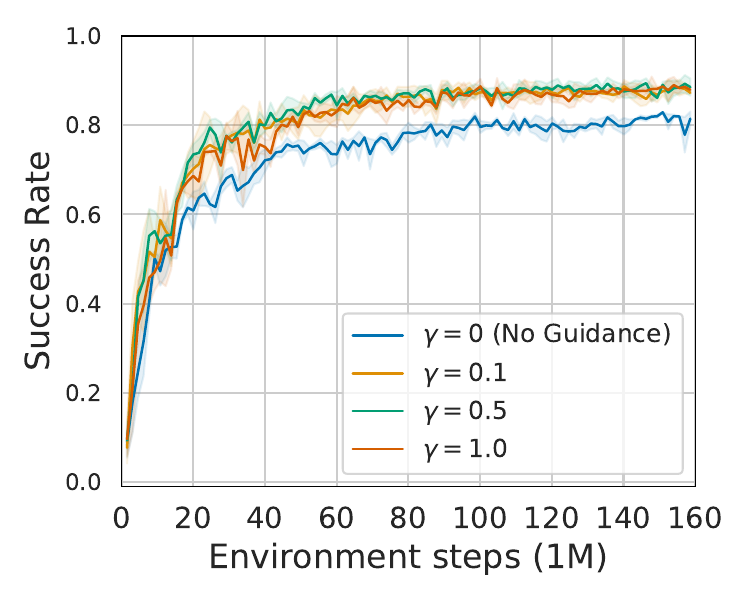}
    \caption{\textbf{Guidance scaling ablation.} We compare the guidance scaling parameter to steer the output of the constraint policy. This result indicates that \ourmethod~without guidance shows worse performance and \ourmethod is robust to a wide range of guidance scaling parameters to achieve better performance.}
    \label{fig:guidance_ablation}
\end{figure}

The degree to which the value function-guided policy coordination improves the task success rate is investigated here. 
Concretely, the impact of the scaling parameter $w$ on the constraint diffusion policy (see Eq.~\ref{eq:guidance}) during teacher policy training is investigated.
As illustrated in Figure~\ref{fig:guidance_ablation}, the teacher policy's performance decreases when value function policy coordination is not applied (\ie $\lambda=0$).
On the other hand, incorporating value function policy coordination consistently enhances the teacher policy's overall performance.
This finding suggests that the constraint policy occasionally generates constraint poses that are suboptimal for the grasping policy. 
Consequently, value function policy coordination promotes on-the-fly adjustments and this is cooperation between the two arms achieves higher success rates.

\section{Experimental Results: Real-World}

In this section, we evaluate a student policy trained using simulated data in real-world environments.
We design experiments to address (1) What is the performance of \ourmethod~when operating over seen and unseen objects in the real-world? (2) Does the performance of \ourmethod~improve when conditioned on a desired grasp pose?

\begin{figure}[t]
    \centering
    \includegraphics[width=0.9\linewidth]{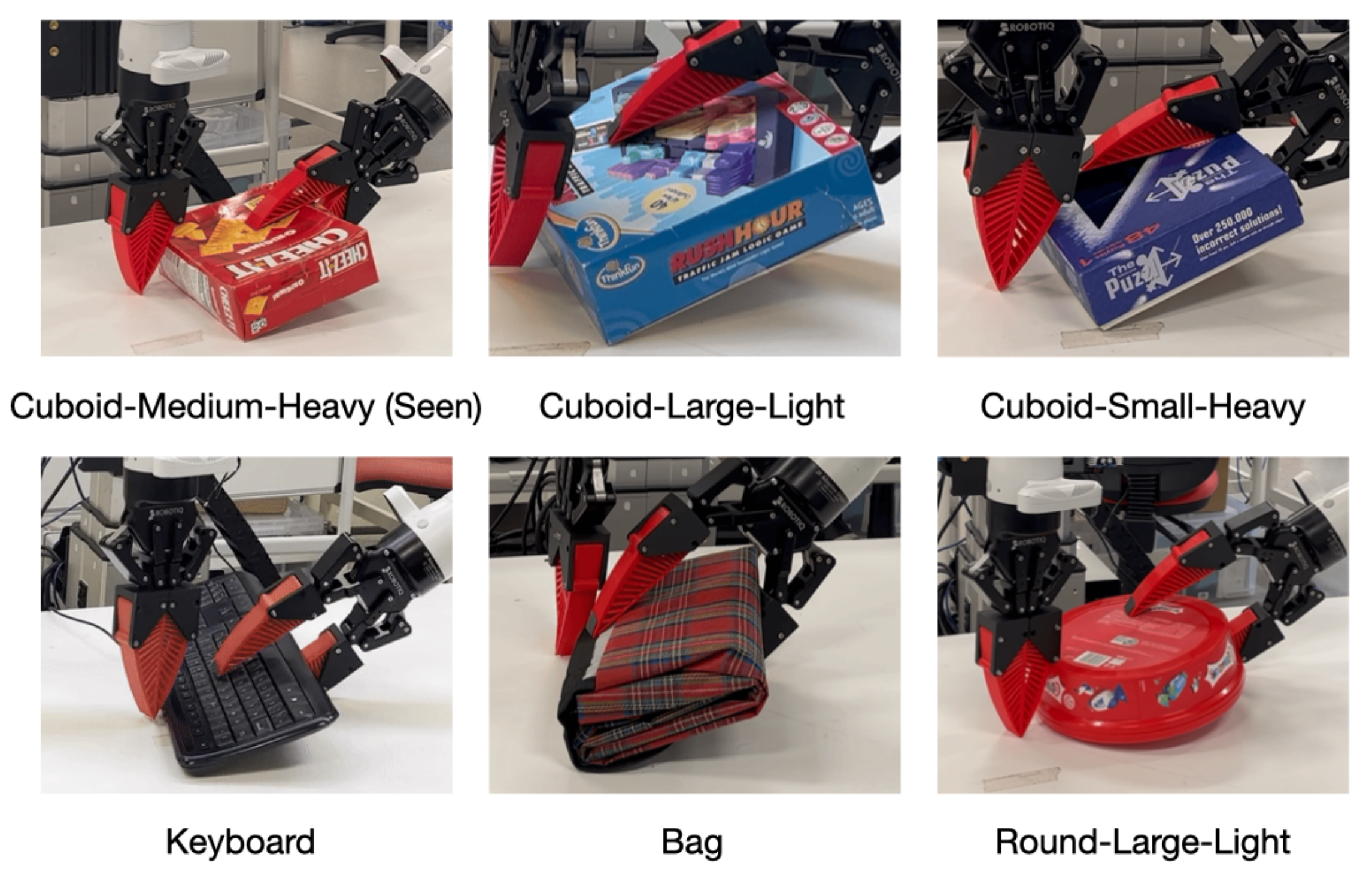}
    \caption{We select several objects with differing sizes and weights that necessitate occluded grasping to evaluate \ourmethod~in the real-world environment.}
    \label{fig:real_objects}
\end{figure}

\begin{table}[t]
    \centering
    \begin{tabular}{c| c | c }
    \toprule
     & \ourmethod & \ourmethod \\
     &  & w/o grasp pose \\
    \hline
    Cuboid-Medium-Heavy (Seen)  & $80\% \ (8/10)$ & $80\% \ (8/10)$ \\
    \rowcolor{gray!20}
    Cuboid-Large-Light  & $90\% \ (9/10)$ & $80\% \ (8/10)$ \\
    Cuboid-Small-Heavy  & $50\% \ (5/10)$ & $60\% (6/10)$ \\
    \rowcolor{gray!20}
    Keyboard  & $80\% \ (8/10)$ & $40\% (4/10)$ \\
    Bag  & $80\% \ (8/10)$ & $80\% \ (8/10)$ \\
    Round-Large-Light  & $30\% \ (3/10)$ & $10\% \ (1/10)$ \\
    \hline
    \rowcolor{gray!20}
    Average & $\mathbf{68.3\% \ (41/60)}$ & $58.3\% \ (35/60)$ \\
    \end{tabular}
    \caption{Performance of \ourmethod~in real-world environments for seen and unseen objects with varying shapes, sizes, and weights.}
    \label{tab:real_world}
\end{table}

\subsection{Experiment Setup}
In this experiment, the student policies are evaluated using both seen and unseen objects with varying shapes, weights, and sizes, as illustrated in Figure~\ref{fig:real_objects}.
To facilitate grasping, we scan the objects to reconstruct their 3D meshes and employ antipodal sampling to generate desired grasp poses. 
This avoids reliance on off-the-shelf grasp pose prediction models~\cite{mousavian20196}, as addressing the inaccuracies of such models is beyond the scope of our evaluation.
Nonetheless, \ourmethod~is compatible with any grasp pose prediction model that takes a segmented object point cloud as input.

When the student policies are conditioned on the desired grasp poses, we estimate the object pose and thus can infer the desired grasp pose during manipulation in real-time, similar to prior work~\cite{wan2023unidexgraspimprovingdexterousgrasping}.
To achieve this, FoundationPose~\cite{wen2024foundationpose} is used to track object pose.
Additioanlly, SAMTrack~\cite{cheng2023segment} is used to generate a segmentation mask to extract the target object point cloud used as input for the policies.

After generating a constraint pose using the student constraint policy, the desired joint positions for the right arm are obtained using the IK solver in CuRobo~\cite{sundaralingam2023curobo}.
Then, MoveIt~\cite{Coleman2014ReducingTB} is used to control the right arm to the desired positions.
After the right arm is positioned in the constraint pose, the left arm begins the student grasping policy rollout.

\subsection{Results}
Table~\ref{tab:real_world} shows the performance of the student policy in the real world.
\ourmethod~effectively tackles occluded grasping tasks for both seen and unseen objects. 
Nonetheless, it encounters challenges with the round box, which is particularly difficult to stabilize in real-world conditions. 
While performance shows a slight decline, \ourmethod~still achieves a reasonable level of success, even without the target grasp pose as input.
\ourmethod, when operating without the target grasp pose, struggles to recover from failed non-prehensile manipulation attempts. 
When an initial push fails to move the object as intended, it is unable to retract and retry the push. 
For instance, pushing a keyboard towards a constraint is particularly challenging due to its thin profile, resulting in a low success rate of only $40\%$, as the left arm often fails to effectively position the keyboard.

In contrast, incorporating the desired grasp pose enables \ourmethod~to recover from such failures by reattempting the task until the left arm successfully reorients and completes the grasp. 
This demonstrates the importance of the desired grasp pose as a critical input for guiding the policy and improving performance.
However, the version of \ourmethod~without the desired grasp pose offers increased flexibility, as it eliminates the need for real-time object pose estimation. 
This trade-off makes it more practical for scenarios where tracking the target grasp pose is infeasible.

\section{Limitations}

\ourmethod~offers notable improvements in learning efficiency and generalisation compared to baselines and prior occluded grasping methods. 
However, there are some limitations to consider.
Firstly, \ourmethod~struggles with unseen objects of significantly different shapes, which could be addressed by training the teacher and student policy with a more diverse set of geometries. 
Additionally, \ourmethod~faces challenges with round objects in the real world, where stabilisation during occluded grasping is difficult. 
This issue could be mitigated through a closed-loop control approach, such as learning a residual policy for real-time constraint pose adjustments.
\section{Conclusion} 
We present \ourmethod, a bimanual robotic system for occluded grasping tasks.
By introducing a constraint policy and value function-guided policy coordination that refine the generated constraint pose using gradients from a value function, we demonstrate that the coordinated policies can efficiently learn to solve challenging occluded grasping tasks.
Furthermore, the trained teacher policies are distilled into student policies without dependence on privileged information for real-world deployment.
Through empirical evaluation, we show that \ourmethod~achieves significantly better performance compared to a state-of-the-art baseline and instantiations of \ourmethod~in both simulated and real-world environments.

\section*{Acknowledgments}
This work was supported by a UKRI/EPSRC Programme Grant [EP/V000748/1]. We would also like to thank the SCAN facility for carrying out this work  (\url{http://dx.doi.org/10.5281/zenodo.22558}).

\bibliographystyle{plainnat}
\bibliography{references}

\clearpage

\appendix
\section{Appendix}
\subsection{Additional Analysis for Experiments}
\label{appendix:analysis}
\subsubsection{Student Policy Performance per Object}
\begin{figure*}
    \centering
    \includegraphics[width=0.95\textwidth]{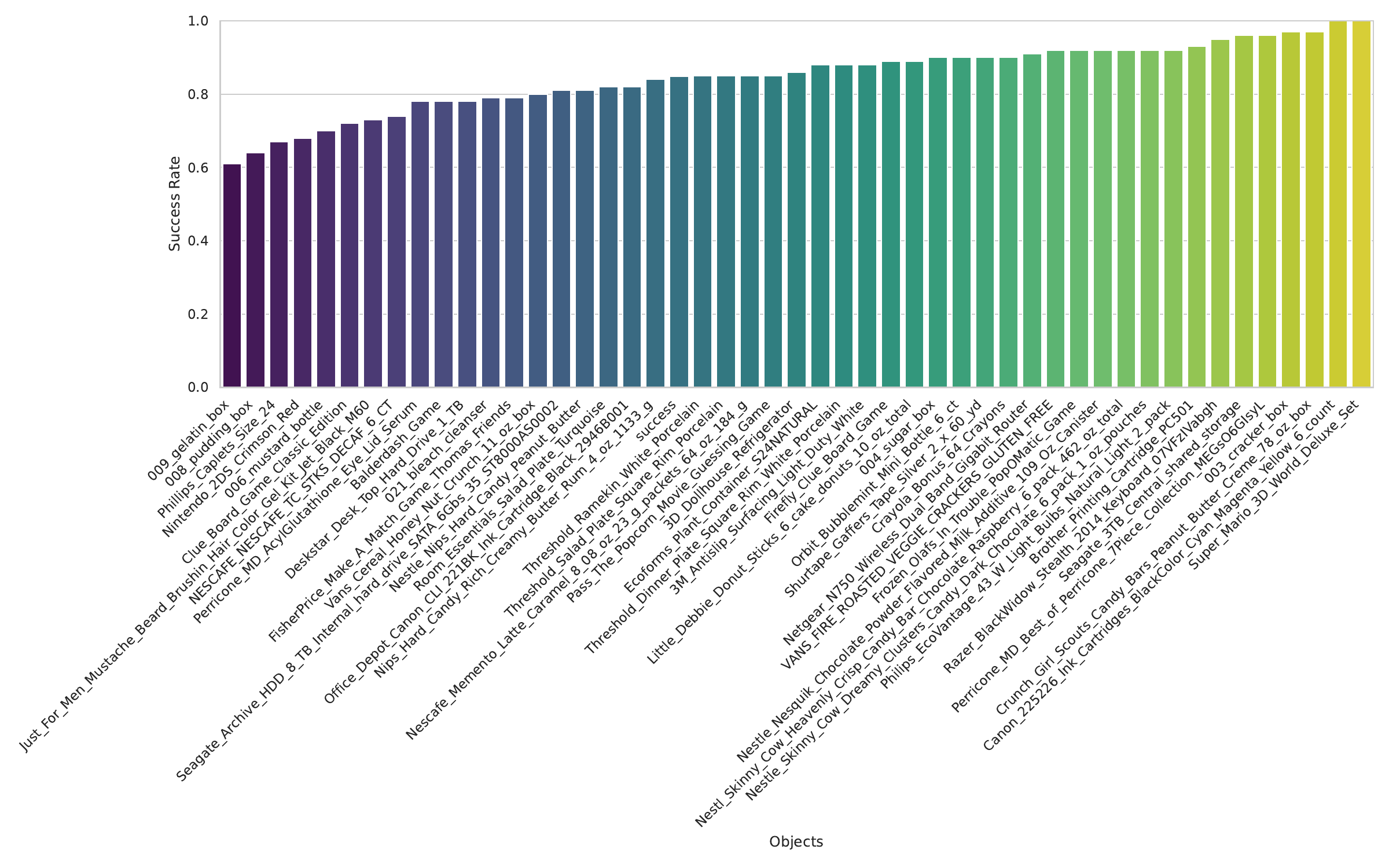}
    \caption{\textbf{Grasp performance of \ourmethod's student policies for each object in simulation. The success rate is averaged over $50$ trials.}}
    \label{fig:student_train_success}
\end{figure*}

Figure~\ref{fig:student_train_success} illustrates the success rate of \ourmethod~for each object used during training.
While \ourmethod~demonstrate performant success rate across diverse objects, the occluded grasp performance for small objects or objects with complex geometries is reduced when compared to that of large objects with simple geometries.
In order to overcome this limitation, it is suggested that both teacher and student policies be trained using more diverse objects, such as those available in the Objaverse datasets~\cite{deitke2023objaverse}.

\subsection{Teacher Policy Details}
\label{appendix:teacher}

\subsubsection{Teacher Constraint Policy}
\label{appendix:teacher_const}
We employ a diffusion policy~\cite{chi2023diffusion} as the basis for the teacher constraint policy. The diffusion policy is implemented using a Denoising Diffusion Probabilistic Model (DDPM), with a multi-layer perceptron (MLP)-based backbone.
The denoising model is built on a three-level UNet architecture, comprising residual blocks with a hidden layer size of $512$.
The diffusion time step is encoded as an $80$-dimensional feature vector. 
Additionally, the desired grasp pose, $\mathbf{x} \in \mathbb{R}^{9}$, and the object's ID, $\mathbf{x}_{obj\_id} \in \mathbb{R}^{16}$, are encoded into an $80$-dimensional vector respectively to provide task-specific context. 
Similarly, the noisy input representing the constraint pose is encoded into another $80$-dimensional vector. 
These encoded vectors are summed and passed through the residual blocks. 
The denoising model outputs the noise added to the original input during the forward diffusion process.
In this work, we use $100$ diffusion time steps for both training and inference.
We train the diffusion policy using an Adam optimiser with a learning rate of $1\times 10^{-4}$.

\subsubsection{Teacher Grasping Policy}
\label{appendix:teacher_manipulation}
We train a teacher grasping policy using Proximal Policy Optimisation (PPO).
An actor network consists of an MLP with $2$ hidden layers of sizes $[256, 256]$.
The actor network is parameterized as a Gaussian distribution with a fixed, state-independent standard deviation.
The critic network consists of an MLP with $3$ hidden layers of sizes $[256, 256, 256]$.

We define the privileged information used to train the policy as $[\mathbf{x}_{\text{robot}}, \mathbf{x}_{\text{goal}}, \mathbf{x}_{\text{obj}}] \in \mathbb{R}^{64}$.
The robot proprioceptive states, $\mathbf{x}_{\text{robot}}$, include the left end-effector pose, $\mathbf{x}_{\text{left}} \in \mathbb{R}^{9}$, the right end-effector pose, $\mathbf{x}_{\text{right}} \in \mathbb{R}^{8}$, and the translational and rotational action scale parameters for the operational space controller, $\mathbf{x}_{\text{control}} \in \mathbb{R}^{2}$.
The right end-effector states, $\mathbf{x}_{right}$, exclude the $z$-coordinate position, as the table height remains constant, and the constraint pose is fixed at a predetermined $z$-coordinate.
The goal-related states, $\mathbf{x}_{\text{goal}}$, consist of the desired grasp pose, $\mathbf{x}_{\text{grasp}} \in \mathbb{R}^{7}$, the distance between the left end-effector and the desired grasp position, $\mathbf{x}_{\text{dist}} \in \mathbb{R}^{3}$, and the orientation distance between the left end-effector and the desired grasp orientation in the axis-angle representation, $\mathbf{x}_{\text{dist\_ori}} \in \mathbb{R}^{3}$.
The object states, $\mathbf{x}_{\text{obj}}$, comprise the object pose, $\mathbf{x}_{\text{obj\_pose}} \in \mathbb{R}^{7}$, the object velocity, $\mathbf{x}_{\text{obj\_vel}} \in \mathbb{R}^{6}$, the friction parameters, $\mathbf{x}_{\text{friction}} \in \mathbb{R}^{2}$, the object's mass, $x_{\text{mass}} \in \mathbb{R}^{1}$, and the object's ID, $\mathbf{x}_{\text{obj\_id}} \in \mathbb{R}^{16}$.

We train the policy using an Adam optimiser with an adaptive learning rate scheduler\footnote{\url{https://skrl.readthedocs.io/en/latest/api/resources/schedulers/kl_adaptive.html}} based on the KL divergence between the current policy and the previous policy, whose maximum learning rate is $1\times 10^{-2}$ and the minimum is $1\times 10^{-6}$.
We use a discount factor of $0.99$, a GAE lambda value of $0.95$, and an entropy coefficient of $6e-3$.
After each policy rollout, the policy is updated using a batch size of $2048$ for $8$ epochs.

\begin{figure}[t]
    \centering
    \includegraphics[width=0.4\textwidth]{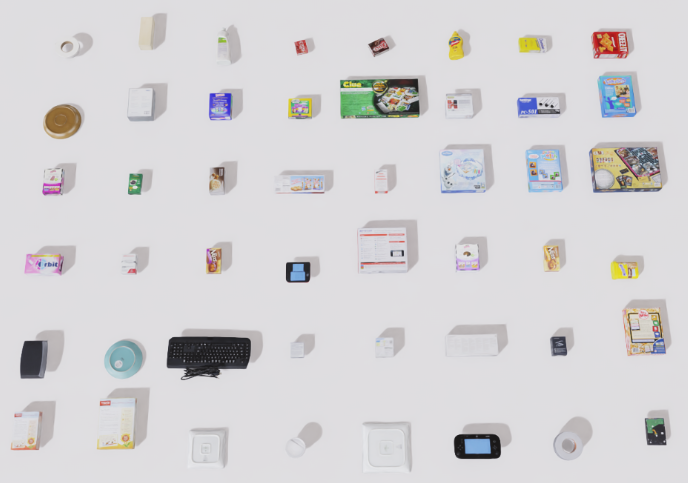}
    \caption{\textbf{Training objects.} We choose $48$ training objects from the Google Scanned Object Dataset~\cite{downs2022googlescannedobjectshighquality}.}
    \label{fig:training_objects}
\end{figure}

\begin{figure}[t]
    \centering
    \includegraphics[width=0.4\textwidth]{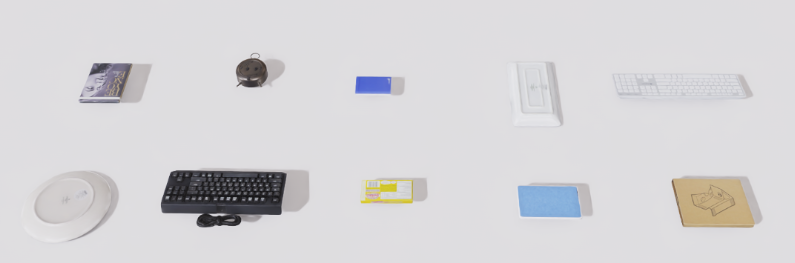}
    \caption{\textbf{Test objects.} We evaluate $10$ held-out objects from the Google Scanned Object Dataset.}
    \label{fig:test_objects}
\end{figure}

\subsubsection{Reward function}
The reward function used in our experiments comprises six terms and is defined as follows:
\begin{equation}
    \begin{split}
    r = \alpha_{1} r_{dist\_pos} &+ 
    \alpha_{2} r_{dist\_ori} - \alpha_{3} r_{collision} \\
    & - \alpha_{4} r_{action} + \alpha_{5} r_{lift} + \alpha_{6} r_{success}        
    \end{split}
\end{equation}
where the weighting coefficients are set to $\alpha_1 = 0.2$, $\alpha_2 = 0.2$, $\alpha_3 = 1.0$, $\alpha_4 = 0.025$, $\alpha_5 = 0.1$, and $\alpha_6 = 40$. Each term in the reward function serves a distinct purpose in guiding the robot’s behaviour:

\begin{itemize}
    \item \textbf{Position Distance Reward ($r_{dist\_pos}$):} 
    This term incentivizes the left end-effector to move towards the desired grasp position. It is computed as:
    \begin{equation}
        r_{\text{dist\_pos}} = 1 - \tanh(4 \cdot ||\mathbf{p}_{\text{left}} - \mathbf{p}_{\text{grasp}}||_{2}),
    \end{equation}
    where $\mathbf{p}_{\text{left}} \in \mathbb{R}^3$ and $\mathbf{p}_{\text{grasp}} \in \mathbb{R}^3$ represent the current and desired positions of the left end-effector, respectively.

    \item \textbf{Orientation Distance Reward ($r_{dist\_ori}$):} 
    This term encourages the left end-effector to align its orientation with the desired grasp orientation. The orientation difference is measured in the axis-angle space`. The reward is computed as:
    \begin{equation}
        r_{\text{dist\_ori}} = 1 - \tanh(0.2 \cdot ||\boldsymbol{\theta}_{\text{left}} - \boldsymbol{\theta}_{\text{grasp}}||_{2}),
    \end{equation}
    where $\boldsymbol{\theta}_{\text{left}} \in \mathbb{R}^3$ and $\boldsymbol{\theta}_{\text{grasp}} \in \mathbb{R}^3$ represent the axis-angle representations of the current and desired orientations of the left end-effector, respectively.

    \item \textbf{Action Penalty ($r_{action}$):} This term discourages large control commands by penalizing the magnitude of the action vector:
    \begin{equation}
        r_{\text{action}} = ||\mathbf{a}||_2.
    \end{equation}

    \item \textbf{Collision Penalty ($r_{\text{collision}}$):} To prevent self-collisions and contact with the table, we compute the signed distance (SD) using CuRobo~\cite{sundaralingam2023curobo}. The collision penalty is given by:
    \begin{equation}
        r_{\text{collision}} = SD_{\text{self\_col}} + SD_{\text{table}}.
    \end{equation}
    The signed distance is computed for the robot arms, excluding the grippers, since the grippers must make contact with the table for occluded grasping problems. In CuRobo, a positive signed distance indicates a collision.

    \item \textbf{Lift Reward ($r_{\text{lift}}$):} This term encourages lifting the object to expose an initially occluded grasp pose. It is defined as an indicator function:
    \begin{equation}
        r_{\text{lift}} = \mathbbm{1}(z_{\text{grasp}} > z_{\text{grasp,init}} + 2 \text{ cm}),
    \end{equation}
    where $z_{\text{grasp}}$ and $z_{\text{grasp,init}}$ denote the current and initial heights of the desired grasp position, respectively.

    \item \textbf{Grasp Success Reward ($r_{success}$):} At the end of an episode, a reward of 1 is assigned if the left arm successfully grasps and lifts the object; otherwise, the reward is 0:
    \begin{equation}
        r_{\text{success}} =
        \begin{cases}
            1, & \text{if grasp and lift are successful}, \\
            0, & \text{otherwise}.
        \end{cases}
    \end{equation}
\end{itemize}

\subsection{Student Policy Details}
\label{appendix:student}

\subsubsection{Studnet Constraint Policy}
The student constraint policy integrates the DP3 encoder~\cite{ze20243d} and a state encoder to process point cloud and state observations, respectively. 

The DP3 encoder comprises three fully connected layers with dimensions of $[128, 256, 384]$, followed by a max pooling operation and a final fully connected layer of size $64$. 
Layer normalization and ReLU activations are applied after each of the initial three layers preceding the max pooling operation.
The state encoder consists of two hidden layers with dimensions of [$128$, $256$]. The state encoder outputs a feature vector of size $32$ given the desired grasp pose $\mathbf{x}_{grasp}$.

The feature vectors produced by the DP3 and state encoders are concatenated and subsequently processed through a MLP to generate a constraint pose.
For this work, the student policy utilizes a Gaussian Mixture Model (GMM)-based approach due to its simplicity and effectiveness. 
Specifically, the GMM-based policy employs $5$ modes, with a minimum standard deviation of $1\times 10^{-4}$.
We employ an AdamW optimiser with a learning rate of $5 \times 10^{-5}$ and a weight decay of $5 \times 10^{-5}$.

\subsubsection{Student Grasping Policy}
We adopt the 3D Diffusion Policy (DP3)~\cite{ze20243d} as the foundation for the student grasping policy. 
The architecture of the DP3 encoder and the state encoder is consistent with that employed in the student constraint policy. 
However, the weights of these encoders are independently initialized from those of the constraint policy. 
Furthermore, the input dimension for the state encoder in the manipulation policy differs from that of the constraint policy.
The state encoder for the manipulation policy processes $\mathbf{x}_{robot}$ and optionally $\mathbf{x}_{grasp}$ as input. 
During training, we employ $100$ diffusion timesteps, whereas during inference a Denoising Diffusion Implicit Model (DDIMs) is used with  $10$ diffusion timesteps to accelerate action generation. 
We use an AdamW optimiser with a learning rate of $5 \times 10^{-5}$ and a weight decay of $5 \times 10^{-5}$.

\subsection{Simulation Setup}
\subsubsection{Training}
\label{appendix:sim_training}
In order to train a teacher policy from a diverse set of objects, we select $48$ objects from the Google Scanned Object dataset, as illustrated in Figure~\ref{fig:training_objects}.
To train teacher policies efficiently, we spawn $1024$ robots and objects in the simulated environment.

In order to train a policy robust to noises and effectively transfer it to real-world environments, we apply domain randomisation during teacher policy training.
Table~\ref{table:domain_randomisation} describes the details of the randomisations used in our experiments.
We also apply domain randomisation during the self-supervised data collection for the constraint policy.

\begin{table}[h]
\centering
\caption{Domain Randomisation Hyperparameters}
\vspace{0.5em}
\begin{tabular}{c|c} 
 \toprule
 Parameter  & Description  \\ 
 \midrule
 \midrule

 Initial robot joint positions & Add noise sampled from $\mathcal{N}(0, 0.05)$  \\
 Robot base position & Add random noise sampled from $\mathcal{U}(-0.015, 0.015)$ \\
 & to the z-coordinate of the robot base \\
 PID position action scale & Sampled from $\mathcal{U}(0.03, 0.04)$ \\
 PID rotation action scale & Sampled from $\mathcal{U}(0.1, 0.2)$ \\
 Action & Add random noise sampled from $\mathcal{N}(0, 0.01)$ \\
 Object mass & Add mass sampled from $\mathcal{U}(-0.1, 0.1)$ \\
 Static and dynamic friction & Sampled from $\mathcal{U}(0.8, 1.2)$ \\
 Grasp position & Add random noise sampled from $\mathcal{N}(0, 0.005)$ \\
 Grasp translational distance & Add random noise sampled from $\mathcal{N}(0, 0.005)$ \\
 Grasp rotational distance & Add random noise sampled from $\mathcal{N}(0, 0.005)$ \\
 End-effector position & Add random noise sampled from $\mathcal{N}(0, 0.01)$ \\
 Object position & Add random noise sampled from $\mathcal{N}(0, 0.01)$ \\
 Object orientation & Add random noise sampled from \\
 & $U(-0.2\pi\: \text{rad}, 0.2\pi\: \text{rad})$ to the yaw axis \\
 \bottomrule
\end{tabular}
\label{table:domain_randomisation}
\end{table}

\subsubsection{Evaluation}
To evaluate policies for both seen and novel objects, we also select $10$ held-out objects from the Google Scanned Object dataset (see Figure~\ref{fig:test_objects}).

\subsection{Real-World Experiment Setup}
\subsubsection{Input Observation for Student Policies}
The distilled student policies take point clouds as input in real-world environments.
We render depth images with the size of $640\times480$ from a Realsense L515 camera to reconstruct point cloud observations.
Similar to \cite{ze20243d}, we crop the point cloud within a pre-defined bounding box such that it includes the robot arms and the target object.
Then, we remove statistical outliers from the point clouds reconstructed from depth images and apply farthest point sampling to sub-sample $1024$ points.

\subsubsection{Desired Occluded Grasp Pose Generation}
In order to scan an object to reconstruct a mesh, we use Polycam, an application that captures pictures of objects and reconstructs an object mesh using Neural Radiance Fields (NeRF).
Using the reconstructed mesh, we generate desired occluded grasp poses using antipodal sampling.

\subsection{Baseline Method Details}

\subsubsection{PPO}
We train a policy using Proximal Policy Optimization (PPO)~\cite{schulman2017proximal}, where the policy outputs 12-dimensional delta end-effector poses corresponding to both the left and right arms. 
We use the same hyperparameters employed for training \ourmethod, except for the entropy coefficient, which is set to $0.003$. 
This modification was made because using the original entropy coefficient caused a continuous increase in the policy's standard deviation, resulting in the policy's inability to exploit a stable and effective strategy during training.

\subsubsection{PPO + Constraint Reward}
Similar to the \emph{PPO} baseline, but we introduce an additional reward term that encourages the right arm to be used as a constraint.
In particular, we add a reward $r_{right\_dist}=||T^{obj} - T^{RightEE}||_{2}$.

\subsubsection{\ourmethod~w/ Fixed Constraint}
Instead of employing a trained constraint policy, we place the right arm as a constraint at a fixed pose.
To accommodate objects of varying sizes and orientations, the constraint is positioned at the right hand side of the workspace rather than at the centre.
This policy is trained using the same hyperparameters as those employed by \ourmethod.

\end{document}